%% file: main.tex
\documentclass{article}

\PassOptionsToPackage{square,numbers}{natbib}

\usepackage[preprint]{neurips_2025}
\usepackage[utf8]{inputenc} % allow utf-8 input
\usepackage[T1]{fontenc}    % use 8-bit T1 fonts
\usepackage{hyperref}       % hyperlinks
\usepackage{url}            % simple URL typesetting
\usepackage{booktabs}       % professional-quality tables
\usepackage{amsfonts}       % blackboard math symbols
\usepackage{nicefrac}       % compact symbols for 1/2, etc.
\usepackage{microtype}      % microtypography
\usepackage{xcolor}         % colors
\usepackage{graphicx}
\usepackage{textcomp}
\usepackage{xcolor}
\usepackage{comment}
\usepackage{balance}
\usepackage{multirow}
\usepackage{makecell}
\usepackage{wrapfig}

\usepackage{amsmath,amssymb,amsfonts}
\usepackage{subfig}

\newcommand{\etal}{{\textit{et al.}}}
\newcommand{\ie}{{\textit{i.e. }}}

\title{Directly Learning Stock Trading Strategies Through Profit Guided Loss Functions}

\author{%
  Devroop Kar \\
  Rochester Institute of Technology \\
  Rochester, NY \\
  \texttt{dk7405@rit.edu}
  \And
  Zimeng Lyu \\
  Rochester Institute of Technology \\
  Rochester, NY \\
  \texttt{zimenglyu@mail.rit.edu} \\
  \And
  Sheeraja Rajakrishnan \\
  Rochester Institute of Technology \\
  Rochester, NY \\
  \texttt{sr8685@rit.edu} \\
  \And
  Alex Ororbia \\
  Rochester Institute of Technology \\
  Rochester, NY \\
  \texttt{agovcs@g.rit.edu} \\
  \AND
  Hao Zhang \\
  Rochester Institute of Technology \\
  Rochester, NY \\
  \texttt{hzhang@saunders.rit.edu} \\
  \And
  Travis Desell \\
  Rochester Institute of Technology \\
  Rochester, NY \\
  \texttt{tjdvse@rit.edu} \\
  \And
  Daniel Krutz \\
  Rochester Institute of Technology \\
  Rochester, NY \\
  \texttt{dxkvse@rit.edu} \\
}

\begin{document}

\maketitle

\begin{abstract}
\input{sections/00--abstract}
\end{abstract}

\input{sections/01--introduction}

\input{sections/02--literature}

\input{sections/03--methodology}

\input{sections/04--analysis}

% \devroop{test year - 2022, 2021
% models ?? -> TSF non-transformer based, RNN/LSTM/GRU networks | Risk mitigation - hold node | number of stocks ---> different portfolios (3)}

\input{sections/05--setup}

\input{sections/06--results}

\input{sections/07--conclusion}

\begin{ack}
This material is based upon work supported by the United States National Science Foundation under grant \#2225354.
\end{ack}

\bibliography{main}

% References follow the acknowledgments in the camera-ready paper. Use unnumbered first-level heading for
% the references. Any choice of citation style is acceptable as long as you are
% consistent. It is permissible to reduce the font size to \verb+small+ (9 point)
% when listing the references.
% Note that the Reference section does not count towards the page limit.
% \medskip

% {
% \small

% [1] Alexander, J.A.\ \& Mozer, M.C.\ (1995) Template-based algorithms for
% connectionist rule extraction. In G.\ Tesauro, D.S.\ Touretzky and T.K.\ Leen
% (eds.), {\it Advances in Neural Information Processing Systems 7},
% pp.\ 609--616. Cambridge, MA: MIT Press.

% [2] Bower, J.M.\ \& Beeman, D.\ (1995) {\it The Book of GENESIS: Exploring
%   Realistic Neural Models with the GEneral NEural SImulation System.}  New York:
% TELOS/Springer--Verlag.

% [3] Hasselmo, M.E., Schnell, E.\ \& Barkai, E.\ (1995) Dynamics of learning and
% recall at excitatory recurrent synapses and cholinergic modulation in rat
% hippocampal region CA3. {\it Journal of Neuroscience} {\bf 15}(7):5249-5262.
% }

%%%%%%%%%%%%%%%%%%%%%%%%%%%%%%%%%%%%%%%%%%%%%%%%%%%%%%%%%%%%

% \newpage
\clearpage
\appendix \label{sec:appendix}
\input{sections/08--appendix}

\end{document}

%% file: sections/00--abstract.tex
Stock trading has always been a challenging task due to the highly volatile nature of the stock market. Making sound trading decisions to generate profit is particularly difficult under such conditions. To address this, we propose four novel loss functions to drive decision-making for a portfolio of stocks. These functions account for the potential profits or losses based with respect to buying or shorting respective stocks, enabling potentially any artificial neural network to directly learn an effective trading strategy. Despite the high volatility in stock market fluctuations over time, training time-series models such as transformers on these loss functions resulted in trading strategies that generated significant profits on a portfolio of $50$ different \textit{S\&P $500$} company stocks as compared to a benchmark reinforcment learning techniques and a baseline \emph{buy and hold} method. As an example, using 2021, 2022 and 2023 as three test periods, the Crossformer model adapted with our best loss function was most consistent, resulting in returns of $51.42\%$, $51.04\%$ and $48.62\%$ respectively. In comparison, the best performing state-of-the-art reinforcement learning methods, PPO and DDPG, only delivered maximum profits of around $41$\%, $2.81\%$ and $41.58\%$ for the same periods. The code is available at \url{https://anonymous.4open.science/r/bandit-stock-trading-58C8/README.md}.

%% file: sections/01--introduction.tex
\section{Introduction}
\label{sec:intro}

Rapid advancements in artificial intelligence (AI) have revolutionized numerous industries, including finance. In stock forecasting and portfolio management, AI techniques have shown potential in enhancing the quality of decision-making, mitigating risks, and optimizing returns. The increasing availability of large-scale financial data, combined with sophisticated computational tools, has provided an opportunity to leverage AI to navigate the complexities of financial markets.

Stock forecasting, the process of predicting future price movements or trends, is a cornerstone of investment strategy. Traditional methods, such as fundamental and technical analysis, often rely on human judgment and linear modeling, which can struggle to account for the dynamic and nonlinear nature of financial markets. AI-driven approaches such as machine learning (ML)~\cite{kumbure2022machine}, deep learning (DL)~\cite{sonkavde2023forecasting}, and natural language processing (NLP)~\cite{jing2021hybrid} have the ability to analyze vast amounts of structured and unstructured data. These methods can uncover hidden patterns and relationships, offering a great edge in prediction accuracy. 
%\travis{would be great if you could add citations for these statements, but we may not have enough room}

%, such as the Markowitz Mean-Variance Model
Portfolio management, another critical aspect of investment strategy, involves the selection and allocation of assets to balance risk and reward according to an investor's objectives. Conventional portfolio optimization methods~\cite{gunjan2023brief}, including statistical models such as autoregressive integrated moving average (ARIMA) and vector auto-regression (VAR) often assume static market conditions and require simplifying assumptions that limit their applicability. AI-based portfolio management systems overcome these limitations by dynamically adapting to changing market conditions, integrating real-time data, and employing techniques such as reinforcement learning (RL) and optimization algorithms to construct robust portfolios. New architectures and techniques are being introduced daily to predict such stock fluctuation, or on a broad scale, for time-series forecasting (TSF). %Taking this into account, this paper investigates the following research questions:

Rather than designing different architectures, we introduce four variations of a loss function that would guide an artificial neural network (ANN) to learn a portfolio management strategy. This is in contrast to RL methods which seek to uncover a policy/strategy or other decision-making strategies, e.g., heuristics or bandits, which operate off time-series forecasts or historical data. These loss functions allow ANNs to directly learn trading strategies specifying which stocks to buy or to short and by how much or if assuming a ``hold'' position would be more beneficial. With this in mind, we investigate the following research questions: 

\hangindent=3.6em %\hangafter=1
\textbf{RQ1.} \emph{Can a directly learned strategy allow any deep learning architecture for time-series forecasting to learn inter-stock patterns and maximize profits?~} - We find that utilizing and applying a custom loss function for training transformer-based models provides better overall strategies for generating profits.

\hangindent=3.6em %\hangafter=1
\textbf{RQ2.} \emph{How would the performance of such a strategy compare across different TSF models as well as RL models?~} - We find that, while the ability of an artificial neural network to identify relations across company stocks is a factor, such a loss-function driven approach yields better performance %than RL methods 
in generating profit as compared to other methods.

% \travis{I think we need to reframe this - we're providing a loss function which directly learns a strategy, as opposed to RL methods which provide a strategy or other decision making strategies (i.e., heuristics or bandits) which operate off time series forecasts or historical data.}

\begin{comment}
This research explores the intersection of AI, stock forecasting, and portfolio management, highlighting the methodologies, challenges, and practical applications of AI-driven strategies. We provide an overview of cutting-edge techniques, such as neural networks, sentiment analysis, and algorithmic trading systems, and discuss their implications for financial decision-making. Additionally, we examine the challenges associated with these methods, including overfitting, data quality issues, and the interpretability of AI models. By bridging the gap between theoretical advancements and practical implementations, this study aims to contribute to the growing body of knowledge in AI-powered finance and inspire future research in this dynamic field.

In the sections that follow, we detail the evolution of AI techniques in finance, analyze their performance in stock forecasting and portfolio management, and propose potential avenues for innovation and improvement. Through this exploration, we aim to demonstrate how AI is reshaping the landscape of financial decision-making, offering both opportunities and challenges for practitioners and researchers alike.
\end{comment}

%% file: sections/02--literature.tex
\section{Related Work}
\label{sec:related_work}

%The intersection of AI and finance has garnered significant attention in academic and professional circles, resulting in a growing body of literature exploring AI-driven methods to extend time-series forecasting to stock forecasting and portfolio management. 

A variety of methods have been applied for time-series forecasting of stock data. Some of the most well-known include traditional statistical models, such as the auto-regressive integrated moving average (ARIMA)~\cite{reddy2019predicting} and general auto-regressive conditional heteroskedasticity (GARCH)~\cite{bollerslev1986generalized}. These approaches are limited given that they often fail to recognize the intricate, non-linear dynamics present in stock data. Moreover, these models rely on certain assumptions, such as data stationarity, and struggle with high levels of volatility~\cite{kumbure2022machine}. AI and machine learning techniques have been adopted to address these limitations due to their ability to process large data volumes and uncover complex patterns ~\cite{leippold2022machine}.

Recurrent neural network (RNN) architectures have been used extensively for stock price prediction. Kamijo~\etal~and Kiani~\etal~\cite{kamijo1990stock}~\cite{kiani2008testing} have showcased their efficacy for this task.  LSTM-RNN models have been extensively employed for predicting stock prices and market trends, as shown in studies by Smith~\etal, Gao~\etal, and Ghosh~\etal~\cite{smith2024stock,gao2018share,ghosh2019stock,zhao2021prediction,yao2018high,pawar2019stock,chen2023research,gupta2022stocknet,lyu2025evolve}. Zhang~\etal~\cite{chen2021mean} investigated the application of long short-term memory (LSTM) networks for predicting stock prices, demonstrating their superior performance compared to traditional time-series models, such as ARIMA. Beyond single architectures, Zhang~\etal~demonstrated the efficacy of integrating deep belief networks (DBNs) with LSTMs for stock price movement prediction~\cite{zhang2021predicting}. Hybrid approaches involving RNNs and convolutional neural networks (CNNs) have also been explored; using a hybrid RNN-CNN architecture, Zhang~\etal~and Yang~\etal~\cite{zhang2018new,yang2024separating} were able to improve prediction accuracy. These methods combine the temporal modeling strengths of RNNs with the feature extraction capabilities of CNNs. Kong \etal \cite{kong2025unlocking} proposed P-sLSTM, which utilized patches to divide the time series into manageable portions -- this allowed the underlying sLSTM \cite{beck2024xlstm} to learn short-term dependencies and then combined them to extract global dependencies.

%However, due to challenges such as vanishing gradients and the inability to retain long-term information faced by RNNs, advanced memory cells, including long short-term memory \cite{schmidhuber1997long} (LSTM), gated recurrent units \cite{cho2014learning} (GRU), and delta-RNN \cite{ororbia2017diff} units have been introduced. 

Transformers were originally developed for neural machine translation~\cite{vaswani2017attention} tasks. However, since then, they have been adapted for multivariate time-series forecasting and have shown great potential in capturing temporal patterns and cross-dimensional relationships~\cite{zhang2022transformer, yanez2024stock}. Wu~\etal~\cite{wu2021autoformer} introduced the Autoformer with a novel auto-correlation mechanism in place of the usual transformer self-attention mechanism. The Crossformer, from Zhang~\etal~\cite{zhang2022crossformer}, utilized dimension-segment-wise (DSW) embeddings and a two-stage attention (TSA) layer to model both inter-temporal and inter-dimension patterns effectively. %PatchTST improved long term predictions by independently processing each channel with a patching mechanism and self-supervised pre-training. This helped optimize model accuracy and resource utilization~\cite{nie2022time}. 
Similarly, DeformTime introduced deformable attention blocks (DABs) to enhance pattern recognition across exogenous variables~\cite{shu2024deformtime} at different temporal granularities. %TSMixer was introduced as a lightweight alternative to resource-heavy transformer models. It utilizes a multi-layer perceptron (MLP)-based approach and incorporates reconciliation heads and gated attention to reduce the computational burden~\cite{vijay2023tsmixer}.
DLinear \cite{zeng2023transformers} uses a fully-connected layer \ie MLP, along the temporal dimension with seasonal-trend
decomposition to for future estimations. TimesNet \cite{wu2022timesnet} uses convolutions to discover multiple periods and capture temporal 2D-variations. These models represent some of the best performing time series forecasting models to date.

While forecasts of stock values are useful, they do not provide a full methodology to manage a portfolio of stocks, \ie, given a budget and a selection of stocks, how much of that budget should be allocated to each stock to buy or short for a given time interval. To address this, reinforcement learning (RL) has gradually emerged as a powerful tool for portfolio optimization and stock prediction~\cite{mohammadshafie2024deep}. By modeling the market as an environment that returns maximum rewards based on the correct trading decision, the stock trader agents are trained as much as possible to follow the market behavior by optimizing financial metrics. An intelligent agent can be obtained through value iteration and policy iteration. Among these, Q-learning methods~\cite{watkins1992q} and actor-critic methods~\cite{konda1999actor} are well-known policy iteration methods in RL. Methods involving multi-armed bandits have also been studied with respect to trading and portfolio management~\cite{kar2024enabling, zhang2022tradebot, ni2023contextual}.

Wu~\etal~\cite{wu2020adaptive} proposed a hybrid model based on RL and a gated recurrent unit (GRU) to extract features necessary for adaptive trading decisions based on time-varying stock market data. Taghian~\etal~\cite{taghian2022learning} proposed a deep Q-learning network for single-model single-asset trading whereas Yang~\etal~\cite{yang2023deep} combined deep RL, transformer layers, and U-Net to learn stock trading strategy on a single stock. Huang~\etal~\cite{huang2024novel} introduced a deep SARSA trading strategy using a bidirectional LSTM-attention module to capture the intricate patterns of the stock market.

Among the actor-critic methods, the deep deterministic policy gradients (DDPG) algorithm~\cite{tan2021reinforcement} is an important approach that concurrently learns a Q-function and a policy. Chen~\etal~\cite{chen2022ddpg} utilized multi-scale time-series features using K-lines in tandem with DDPG to develop a trading strategy. Yang~\etal~\cite{yang2020deep} introduced an ensemble strategy using A2C~\cite{mnih2016asynchronous}, DDPG, and PPO for portfolio allocation. Huang~\etal~\cite{huang2024improving} proposed a hybrid technique using imitation learning and A2C for single asset trading while Liu~\etal~\cite{liu2020finrl, liu2021finrl, liu2022finrl} introduced an open source framework termed FinRL that allows users to apply DRL algorithms to problems in quantitative finance~\cite{liu2024dynamic, xiong2018practical, yang2020deep}. The framework supports many actor-critic methods including PPO, DDPG, TD3~\cite{fujimoto2018addressing}, SAC~\cite{haarnoja2018soft}, etc. Proximal policy optimization (PPO) is considered to be the state-of-the-art with regard to actor-critic methods in RL~\cite{schulman2017proximal, pricope2021deep}. Notably, Zhou~\etal~\cite{zou2024novel} employed a simulated environment for trading and carried out a comparative analysis using three actor-critic algorithms: PPO, advantage actor critic (A2C), and DDPG. Beyond this, Lin and Beling~\cite{lin2021end} proposed an end-to-end framework for trading using PPO.

%% file: sections/03--methodology.tex
\section{Methodology}

%The objective of this work is to devise a model-independent approach that simply involved changing the last layer of any neural network and coupled with the proposed loss function would drive the model to maximize profits while searching for correlations among the stock data \travis{to directly learn how much of each stock to by or short, and how much money to not invest (hold) in the case of uncertainty}

\begin{figure*}
    \centering
    \includegraphics[scale=0.38]{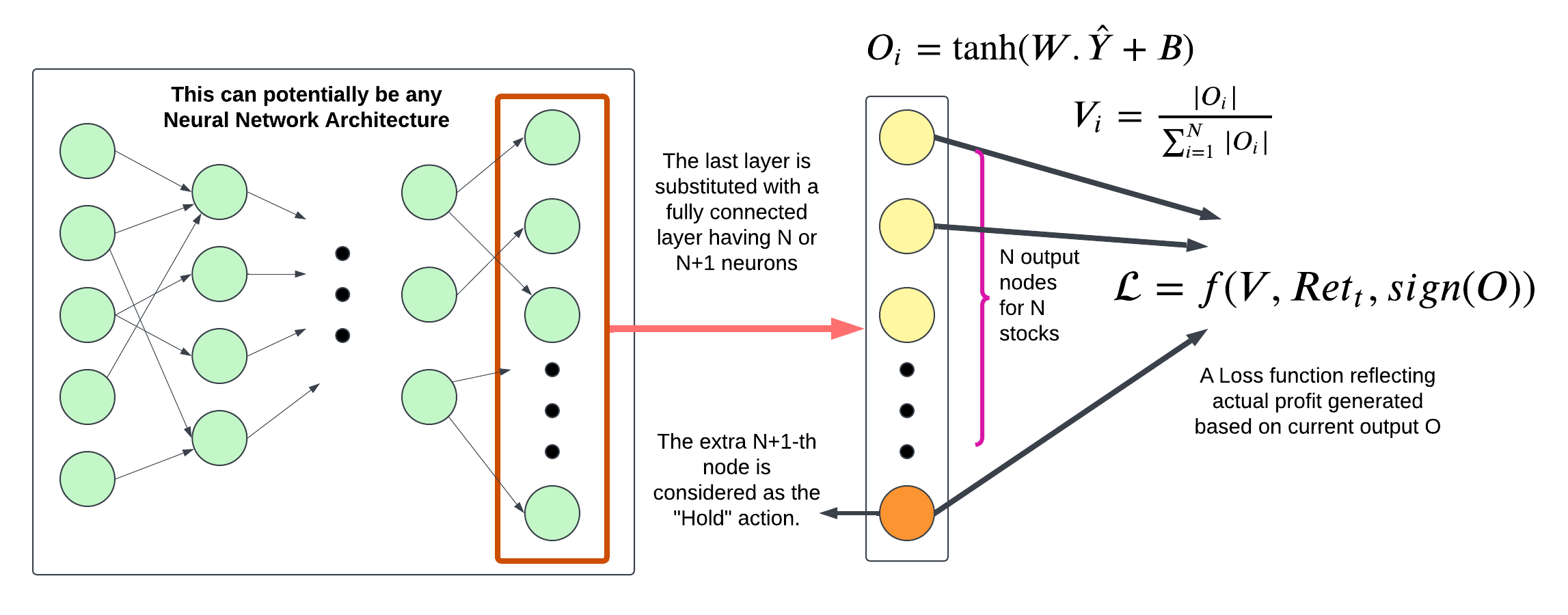}
    \caption{Modifying the final layer of any neural network to capture trading decisions for the present stock portfolio entails using an activation function that bounds outputs between $(-1, 1)$, such as the hyperbolic tangent $\tanh()$. By leveraging one of our new loss functions in tandem with function such as $\tanh()$, the network can then be updated to make daily decisions on whether to buy a stock or short.}
    \label{fig:loss-method}
    \vspace{-0.4cm}
\end{figure*}

The objective of this work is to allow ANNs to directly learn how much of each stock to buy or to short as well as how much money to not invest (hold) in the case of uncertainty to mitigate risk. For this, four different types of loss functions were designed with the goal of having an ANN learn appropriate strategies to maximize profit. Irrespective of the architecture of the ANN, the output layer is designed as a layer of $N+1$ neurons where $N$ is the number of stocks, along with an additional neuron to signify the ``hold'' decision. An activation function such as $\tanh$ is applied to bound the final outputs between the values of $-1$ and $1$. These bounded outputs are then fed into a loss function to directly learn the trading strategy (see Figure \ref{fig:loss-method}).

Considering $O_i = tanh(W^{L} + b^L)$ (where $L$ indexes the final/output layer of a deep ANN) as the final output value for each of the output neurons, we calculate $\hat{V_i} = \frac{|O_i|}{\sum_{i=j}^N|O_j|}$. Given our available funds, $A$, this is the proportion of $A$ we want to spend on a stock to buy or short, i.e., $A * \hat{V_i}$. The exact decision depends on the signum function, i.e., $sign(O_i)$, which can return either $-1$ (in which case we short) or $1$ (in this case we buy). With the \texttt{Hold} option, the corresponding proportion of funds is kept aside without being spent. This additional \texttt{Hold} node also serves to learn and minimize risk.

The proposed loss functions we designed given the above setup were conceived based on the net return across different stock portfolios. Designing the loss functions to reflect future returns allows the ANN to search for strategies where the final profit would be maximized. 

\subsection{Loss I: StockLoss}

The first loss that we designed and studied was formally:
\begin{equation}
    \mathcal{L} = -(\sum_{i=1}^{N}\hat{V_i}\cdot (Ret_{i,t+1} - Ret_{i,t})\cdot sign(O_i)
     + \mathbb{I}(e \rightarrow 1)\mathcal{H}(V))
    \label{eq:loss1}
\end{equation}
where $V_i=\frac{|O_i|}{\sum_{j=1}^N|O_j|}$ is the proportion of stocks to be invested based on the output of the ANN. The exact number of shares will depend on a starting budget $\mathcal{B}$. The expression $(Ret_{i,t+1} - Ret_t)$ is the difference between the returns at time steps $t+1$ and $t$ for a stock $i$. This difference motivates the artificial neural network to invest more in stocks that have higher forecasted returns. The sign of the output of the ANN \textit{i.e.} $sign(O_i)$ gives the decision as to whether to buy a stock or to short it. 
The product of all of the above terms are a reflection of the possible profit that can be made for a stock over a time period $T$. So from a trading perspective, the goal is to maximize $\sum_{i=1}^{N}\hat{V_i}\cdot (Ret_{i,t+1} - Ret_{i,t})\cdot sign(O_i)$. However, since ANNs are typically trained to minimize a loss function, a negation is applied to this sum; this further ensures that the network always attempts to maximize the profit margin (as it minimizes our loss). The expression $\mathcal{H}(V)$ denotes an additional output node which can be used (or not) based on the value of a variable $e$ which the user can provide. So $\mathbb{I}(e \rightarrow 1)$ serves as a check to consider this additional option. This node is used to represent the \texttt{Hold} action. The value for this node represents how much of the budget at each time step is kept aside and not used to buy or short any stock. This node thus usefully serves as a "\textit{risk}" mitigator.

While Equation \ref{eq:loss1} encompasses all of the expressions needed to maximize profit, experiments proved that the policies generated by this loss function were quite volatile. As a result, we next considered and studied a normalized version which helps to stabilize the data fluctuations across stocks and allows the network to explore more evenly.

\subsection{Loss II: StockLoss-Max}

The next loss we considered was as follows: 
\begin{equation}
\begin{split}
    \mathcal{L} &= 1 - \sum_{i=1}^{N}\hat{V_i}\cdot\frac{Ret_{i,t+1} - Ret_{i,t}}{\max_{j \in N}(Ret_{j,t+1} - Ret_{j,t})}\cdot sign(O_i) \\
    &- \mathbb{I}(e \rightarrow 1)\mathcal{H}(V)
\end{split}
\label{eq:loss2}
\end{equation}
where we see that the expression $(Ret_{i,t+1} - Ret_t)$ is normalized by the maximum difference of the returns over all stocks in that time window. This ensured that the entire product $\hat{V_i}\cdot\frac{Ret_{i,t+1} - Ret_{i,t}}{\max_{j \in N}(Ret_{j,t+1} - Ret_{j,t})}\cdot sign(O_i)$ would be $<1$. Subtracting this from Equation~\ref{eq:loss1} drove the networks to maximize the profit margins as before, and this normalization stabilized the policies to some degree.

\subsection{LoST III: StockLoss-L2}

As an alternative to the above loss function, we considered the following for introducing a form of normalization to reduce volatility/improve stability of our neural models:
\begin{equation}
\begin{split}
    \mathcal{L} &= 1 - \\
    &\sqrt{\sum_{i=1}^{N}\hat{V_i}\Biggl(\frac{Ret_{i,t+1} - Ret_{i,t}}{\max_{j \in N}Ret_{j,t+1} - Ret_{j,t}}\Biggr)^2+\mathbb{I}(e \rightarrow 1)\mathcal{H}(V)^2} . 
\end{split} 
    \label{eq:loss3}
\end{equation}
Here we see another perspective on normalizing the return terms and using a L2-norm based approach of squaring the individual stock expressions. The effect was similar to the loss function in Equation \ref{eq:loss1}, where minimizing the entire loss function maximized the profit margins.

\subsection{Loss IV: StockLoss-Norm}

The next loss we developed and examined was:
\begin{equation}
\begin{split}
    \mathcal{L} &= 1 - \sum_{i=1}^{N}\frac{|O_i|\cdot(Ret_{i,t+1} - Ret_{i,t})\cdot sign(O_i)}{\sum_{j=1}^N|O_j|\cdot (Ret_{j,t+1} - Ret_{j,t})} \\
    &- \mathbb{I}(e \rightarrow 1)\mathcal{H}(V)
\end{split}
\label{eq:loss4}
\end{equation}
where we observe that we normalize the product of $\hat{V_i}$ and the returns as opposed to just the returns. 
% A further change was introduced into the loss functions for comparison on top of the smoothness. 

\subsection{Price vs. Return Loss Functions}
As variations of the previous four loss functions, rather than just using $Ret_t$ in the loss functions as a measure of profit, $PRC_t$ was used as a substitute \ie $(PRC_{i,t+1} - PRC_{i,t})$ in place of $(Ret_{i,t+1} - Ret_{i,t})$. This made the loss functions a more direct measure of the actual profit. 
% \travis{i would describe these above in the methodology as other options for loss functions, so you can have version 1 through 4 with RET and 1 through 4 with PRC.}

% \begin{enumerate}
%     \item 
%     \begin{equation}
%     \mathcal{L} = -(\sum_{i=1}^{N}\hat{V_i}\cdot (PRC_{i,t+1} - PRC_{i,t})\cdot sign(O_i)
%      + \mathbb{I}(e \rightarrow 1)\mathcal{H}(V))
%     \label{eq:loss1}
% \end{equation}

%     \item 
%     \begin{equation}
%         \begin{split}
%             \mathcal{L} &= 1 - \\
%             &\quad \sqrt{\sum_{i=1}^{N}\hat{V_i}\Biggl(\frac{PRC_{i,t+1} - PRC_{i,t}}{\max_{j \in N}PRC_{j,t+1} - PRC_{j,t}}\Biggr)^2 \mathbb{I}(e \rightarrow 1)\mathcal{H}(V)^2}
%         \end{split}
%         \label{eq:loss3}
%     \end{equation}
%     \item
%     \begin{equation}
%         \begin{split}
%             \mathcal{L} &= 1 - \\
%             &\sqrt{\sum_{i=1}^{N}\hat{V_i}\Biggl(\frac{PRC_{i,t+1} - PRC_{i,t}}{\max_{j \in N}PRC_{j,t+1} - PRC_{j,t}}\Biggr)^2+\mathbb{I}(e \rightarrow 1)\mathcal{H}(V)^2}
%         \end{split}
%         \label{eq:loss3}
%     \end{equation}
%     \item 
%     \begin{equation}
%         \begin{split}
%             \mathcal{L} &= 1 - \sum_{i=1}^{N}\frac{|O_i|\cdot(PRC_{i,t+1} - PRC_{i,t})\cdot sign(O_i)}{\sum_{j=1}^N|O_j|\cdot (PRC_{j,t+1} - PRC_{j,t})} \\
%             &- \mathbb{I}(e \rightarrow 1)\mathcal{H}(V)
%         \end{split}
%         \label{eq:loss4-prc}
%     \end{equation}
% \end{enumerate}

%% file: sections/04--analysis.tex
\section{Analysis}

\begin{figure}[!ht] 
    \centering
    \subfloat[Variation of Loss \ref{eq:loss1} and gradient against an output $O_i$.Demonstrates a overall linear relationship between each output node value and the loss value; also highlights the discontinuity in the gradient.]{%
        \includegraphics[scale=0.4]{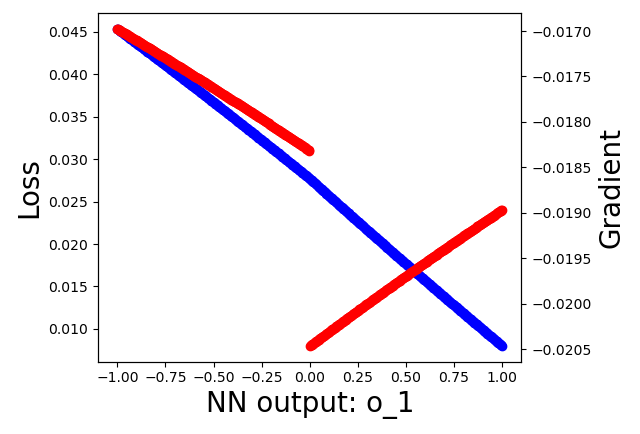}%
        \label{fig:stock-loss-fn-grad}%
        }%
    \hfill%
    \subfloat[Variation of Smooth Loss \ref{eq:loss1} and gradient against an output $O_i$. With the smooth approximation the linear relationship mostly holds, but the discontinuity in the gradient is resolved.]{%
        \includegraphics[scale=0.4]{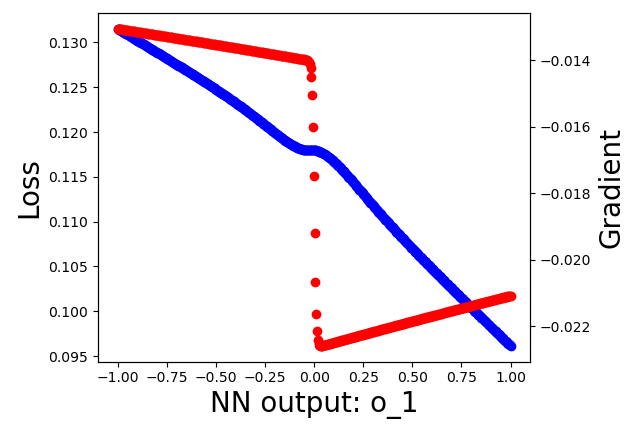}%
        \label{fig:stock-loss-smooth-fn-grad}%
        }%
    \caption{}
\end{figure}

Upon analysis of these loss functions, we identified a possible cause of network training instability. The value of the loss function was plotted against the values of a single output node $O_i$ and its corresponding derivative was also plotted. The output value was varied from $-1$ to $1$ since the $tanh$ function is bounded in that range; only the output node value was varied based on an unchanging return difference. 
Taking loss function \ref{eq:loss1} as the point of discussion, the plots for the function itself and its derivative are shown in Figure \ref{fig:stock-loss-fn-grad}.

% \begin{figure}[!h]
%     \centering
%     \includegraphics[scale=0.45]{images/Stock_loss_gradient_st50_long.png}
%     \caption{Variation of Loss \ref{eq:loss1} and gradient against an output $O_i$.Demonstrates a overall linear relationship between each output node value and the loss value; also highlights the discontinuity in the gradient.}
%     \label{fig:stock-loss-fn-grad}
% \end{figure}

%\travis{nature? do you mean derivative?}
If we plot the loss function against the values of a single output node, a similar pattern can be observed. Considering Equation \ref{eq:loss1} and ignoring the \texttt{Hold} option for simplicity, we examine:
$$\mathcal{L} = -\sum_{i=1}^{N}\hat{V_i}\cdot(Ret_{i,t+1} - Ret_{i,t})\cdot sign(O_i).$$
If we consider the return terms and the values of the other output nodes barring $O_k$ as constant, their sum is also constant \ie $C_1$, and the equation can be rewritten as follows:
\begin{align*}
   \mathcal{L} &= -\sum_{i=1,i\neq k}^{N}\hat{V_i}\cdot(Ret_{i,t+1} - Ret_{i,t})\cdot \text{sign}(O_i) \\
   &\quad - V_k(Ret_{k,t+1} - Ret_{k,t})\cdot \text{sign}(O_k)
\end{align*}

$$\Rightarrow \mathcal{L} = C_1 - \frac{|O_k|\cdot sign(O_k)}{\sum_{j\in N}|O_j|}(Ret_{k,t+1} - Ret_{k,t})$$

$$\Rightarrow \mathcal{L} = C_1 - O_k\cdot\frac{(Ret_{k,t+1} - Ret_{k,t})}{\sum_{j\in N}|O_j|}$$
For a large number of stocks, the sum $\sum_{j\in N}|O_j|$ can be approximated as a constant \ie $C_2$ against a single term as below:
\begin{equation}
\label{eq:loss1_approx}
\Rightarrow \mathcal{L} \approx C_1 - C_2\cdot O_k . 
\end{equation}
 Using this approximation, the loss function \ref{eq:loss1} behaves linearly with a negative slope when $(Ret_{i,t+1} - Ret_{i,t}) > 0$. This scenario represents when one should buy and sell in day trading. Figure \ref{fig:stock-loss-fn-grad} follows this relationship as well. For $(Ret_{i,t+1} - Ret_{i,t}) < 0$, \ie when one needs to short, the nature of the loss (Equation \ref{eq:loss1_approx}) would indicate the reverse.

In case of the function's derivative or gradient, we obtain the following:
$$
\frac{\partial \mathcal{L}}{\partial O_k} = -\sum_{i=1}^N(Ret_{i,t+1} - Ret_{i,t})\frac{\partial}{\partial O_k}\Biggl(\frac{|O_i|\cdot sign(O_i)}{\sum_{j \in N}|O_j|}\Biggr)
$$

$$
\Rightarrow \frac{\partial \mathcal{L}}{\partial O_k} = -\sum_{i=1}^N(Ret_{i,t+1} - Ret_{i,t})\frac{\partial}{\partial O_k}\Biggl(\frac{O_i}{\sum_{j \in N}|O_j|}\Biggr)
$$

\begin{align*}
\Rightarrow \frac{\partial \mathcal{L}}{\partial O_k} &= -\sum_{i=1}^N(Ret_{i,t+1} - Ret_{i,t})\cdot \\
&\quad \Biggl(\frac{\sum_{j \in N}|O_j|\cdot \frac{\partial O_i}{\partial O_k}-O_i\cdot \frac{\partial}{\partial O_k}(\sum_{j \in N}|O_j|)}{(\sum_{j \in N}|O_j|)^2}\Biggr) 
\end{align*}

\begin{align*}
\Rightarrow \frac{\partial \mathcal{L}}{\partial O_k} &= -\sum_{i=1,i \neq k}^N(Ret_{i,t+1} - Ret_{i,t})\frac{-O_i\cdot sign(O_k)}{(\sum_{j \in N}|O_j|)^2}\\
 &\quad - (Ret_{k,t+1} - Ret_{k,t})\frac{\sum_{j \in N}|O_j|-O_k\cdot sign(O_k)}{(\sum_{j \in N}|O_j|)^2} . 
\end{align*}
Here we observe that the relationship between the output node and the gradient of the loss function is more complex as compared to just the loss. However, the objective is to observe that there is a discontinuity in the gradient at $0$ and thus the gradient is not smooth. This is a consequence of using the $sign(O_i)$ function, which is, strictly speaking, not defined at $0$. Therefore, the neural network cannot learn effectively using the loss function as-is.

\begin{wrapfigure}{r}{0.5\textwidth}
\vspace{-0.2cm}
\includegraphics[scale=0.4]{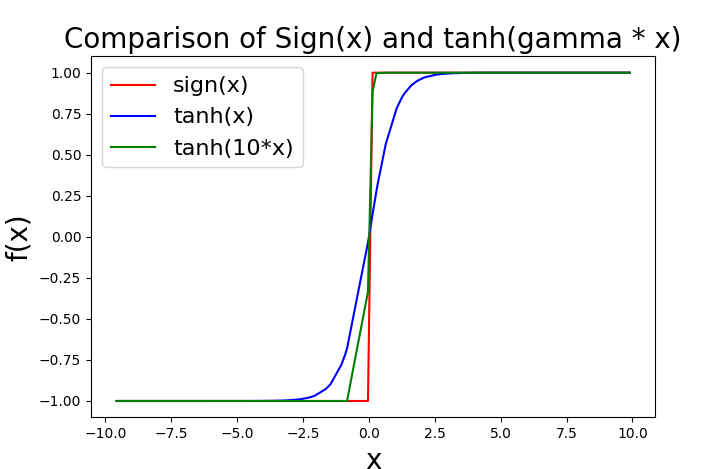}
\vspace{-0.25cm}
    \caption{Comparison between $sign(x)$ and $\tanh(\gamma x)$ for different $\gamma$.}
    \label{fig:func}
\end{wrapfigure}

To overcome the above discontinuity issue, we utilized a smoothing operation to approximate $sign(x)$. Specifically, we employed the sigmoidal function $\tanh(\gamma x)$ in place of $sign(x)$ as the $\tanh()$ function is smooth and defined at $0$ (and $\gamma$ is a coefficient that controls the sharpness/steepness of the sigmoidal funciton's slope). Furthermore, by increasing the value of $\gamma$, the function can be made to closely follow the signum function. An example of this proxy function is shown in Figure \ref{fig:func}; $\gamma$ was taken as $10$ for the experiments of this work.

% \begin{figure}[!h]
%     \centering
%     \includegraphics[scale=0.45]{images/func.png}
%     \caption{Comparison of $sign(x)$ and $\tanh(\gamma x)$ for different values of $\gamma$.}
%     \label{fig:func}
% \end{figure}

The proposed smoothing operation was applied to all of the loss functions we discussed in the prior section. We found that using this soft proxy function, the training process was more stable/less volatile but since this is an approximation, the stability was gained at the cost of some profit margin. A representation of the smoothed version of the loss (Equation \ref{eq:loss1}) and its gradient is shown in Figure \ref{fig:stock-loss-smooth-fn-grad}. %\\

% \begin{figure}[!h]
%     \centering
%     \includegraphics[scale=0.45]{images/Stock_soft_loss_gradient_st50_long.png}
%     \caption{Variation of Smooth Loss \ref{eq:loss1} and gradient against an output $O_i$. With the smooth approximation the linear relationship mostly holds, but the discontinuity in the gradient is resolved.}
%     \label{fig:stock-loss-smooth-fn-grad}
% \end{figure}

% \begin{figure}[!t]
%     \centering
%     \includegraphics[scale=0.3]{images/Stock_loss_gradient_st50_short.png}
%     \caption{Caption}
%     \label{fig:enter-label}
% \end{figure}

%% file: sections/05--setup.tex
\section{Experimental Setup}

\paragraph{Dataset Description} We conducted experiments using daily stock data from the past $30$ years for $50$ S\&P $500$ companies. Overall $8$ numerical festures were used for each stock - stock price, return, bid-ask, spread, turn over, volume change, illiquidity, shares outstanding, and market capitalization. The entire breadth of data was split into train, validation, and test sets. Three partitions were taken with $2021$, $2022$ and $2023$ serving as different testing years. The year before the testing period was the validation period and everything before that formed the training partition. Further details for the datasets used for this work is available in the Appendix. % Data before 2022 was included in the training set, the year 2022 was taken as the validation set, and the year 2023 formed the test set. The stock features used for this work have been explained in Table \ref{tab:financial_metrics}.
%\travis{All model were implemented using Pytorch, adapting the source code from their original projects to include the new loss functions}.
\paragraph{Training Configuration} All models were implemented using Pytorch, adapting the source code from their original projects to include the new loss functions. The models were trained on an \textit{Intel(R) Xeon} server node with 64 GB VRAM and an \textit{NVIDIA A100 GPU}. 

\paragraph{Training Hyperparameters} This work evaluates our loss functions using three transformer Models: DeformTime \cite{shu2024deformtime}, Crossformer \cite{zhang2022crossformer}, and Autoformer \cite{wu2021autoformer} %\travis{give a citation for each}
, a multi-layer perceptron model (DLinear \cite{zeng2023transformers}), a convolutional model (TimesNet \cite{wu2022timesnet}), and a recurrent memory cell model (P-sLSTM \cite{kong2025unlocking}). These models were selected across a wide range of state-of-the art-time series forecasting models to adapt for this work.  These models were also compared to five reinforcement learning (RL) strategies -- DDPG~\cite{tan2021reinforcement}, PPO~\cite{schulman2017proximal}, SAC ~\cite{haarnoja2018soft}, A2C~\cite{mnih2016asynchronous} and TD3~\cite{fujimoto2018addressing}
-- to serve as baselines for the portfolio management task. For the RL strategies, the corresponding strategies used the library FinRL~\cite{liu2020finrl}, where the strategies were fitted for stock trading. This library depends on a specific set of features involving only the opening, closing, adjusted closing stock prices, and volumes exchanged. To overcome this dependency, adjustments were made that allowed the RL models to train on the full set of features stated previously. They were trained using the default configuration parameters found in the \textit{FinRL} library.  Input sequence lengths of $96$ were used to train the time series forecasting models adapted with our loss function. The initial token length was $label\_len=0$ and the prediction sequence length was set at $1$ for all the models. Each model was trained for $100$ epochs with an initial learning rate of $0.001$ and a batch size of $32$. $10$ repeated training runs were conducted for each model and all results shown were generated by selecting the model which performed best on the validation data, which was the data from the year prior to the test period. 

%% file: sections/06--results.tex
\section{Results and Observations}

\begin{table*}
    \centering
    \caption{Comparison between smooth (S) and non-smooth (NS) loss functions ($RET_t$) profit results for 2023}
    \begin{tabular}{lccccc|cccccl}
    \toprule
    \multirow{3}{*}{\textbf{\makecell{Input Seq.\\ Length}}} & \multirow{3}{*}{\textbf{Model}} & \multicolumn{8}{c}{\textbf{Profit \%}} \\
    \cmidrule(lr){3-10}
    & & \multicolumn{4}{c}{\textbf{Hold}} & \multicolumn{4}{c}{\textbf{No Hold}}\\
    \cmidrule(lr){3-10}
    & & \textbf{L-1} & \textbf{L-2} & \textbf{L-3} & \textbf{L-4} & \textbf{L-1} & \textbf{L-2} & \textbf{L-3} & \textbf{L-4} \\
    \midrule
    
    \multirow{12}{*}{\textbf{\textit{96}}} & 
    \textbf{\textit{DeformTime} (NS)} & -5.17 & 0.0 & 5.37 & 1.92 & 3.05 & -2.76 & 15.84 & -2.05 \\
    & \textbf{\textit{DeformTime} (S)} & 0 & 0 & 38.98 & 5.19 & 7.13 & 9.36 & 21.57 & 3.15 \\   
    \cmidrule(lr){2-10}
    
    & \textbf{\textit{Crossformer} (NS)} & 0.43 & 0.66 & 3.07 & 2.78 & 4.33 & -0.59 & -1.26 & 7.09\\
    & \textbf{\textit{Crossformer} (S)} & 2.57 & 0.0 & 38.61 & 6.59 & 7.87 & -1.73 & -0.05 & 9.19 \\
     \cmidrule(lr){2-10}
     
    & \textbf{\textit{AutoFormer} (NS)} & 1.10 & -0.28 & -0.65 & -0.21 & 8.08 & -0.64 & -0.45 & 3.50 \\
    & \textbf{\textit{AutoFormer} (S)} & 0.07 & 0.0 & 1.62 & 0.14 & 0.15 & 1.52 & 0.42 & 0.37 \\
    \cmidrule(lr){2-10}
    
    & \textbf{\textit{TimesNet (NS)}} & 0.36 & -0.41 & 2.56 & 6.37 & 0.58 & 3.37 & 6.38 & 5.63 \\
    & \textbf{\textit{TimesNet (S)}} & 0.19 & 0.0 & 2.79 & 8.85 & 6.18 & 5.54 & 17.27 & 2.40 \\
    \cmidrule(lr){2-10}
    
    & \textbf{\textit{DLinear} (NS)} & 2.03 & 0.0 & 3.28  & 4.96 & 4.25 & 4.93 & 27.60 & 0.96 \\
    & \textbf{\textit{DLinear} (S)} & 5.15 & 0.0 & 14.63 & 3.99 & 9.73 & 6.90 & 10.28 & 2.92 \\
    \cmidrule(lr){2-10}

    & \textbf{\textit{P-sLSTM (NS)}}& 13.75 & -2.90 & 12.72 & -2.86 & 7.77 & -0.43 & 10.44 & 4.08   \\
    & \textbf{\textit{P-sLSTM (S)}} & 23.45 & 6.28 & 25.40 & 15.36 & 11.52 & -0.46 & 15.37 & 5.45 \\

    \bottomrule
    \end{tabular}
    \label{tab:smooth-nosmooth-2023}
    \vspace{-0.4cm}
\end{table*}

\begin{table}[!]
    \centering
    \caption{Mann-Whitney U-Test $p-$values between Smooth and Non-Smooth Loss L3 at confidence level $\alpha=0.05$}
    \begin{tabular}{lcccccccl}
        \toprule
         & \textbf{DeformTime} & \textbf{Crossformer} & \textbf{Autoformer} & \textbf{TimesNet} & \textbf{DLinear} & \textbf{P-sLSTM}\\
        \midrule
        $p-$value & 0.04 & 0.11 & 0.19 & 0.04 & 0.04 & 0.03 \\
         \bottomrule
    \end{tabular}
    \label{tab:utest-smooth}
\end{table}

Experiments were first run to compare the original with the smoothed formulations of the loss functions for the 2023 testing period, with the hold output option turned on and off (see Table \ref{tab:smooth-nosmooth-2023}). Table~\ref{tab:utest-smooth} presents showing that for most models the smooth function is a statistically significant improvement. The Appendix also provides results showing that interestingly the inclusion of the hold node to mitigate risk did not provide a statistically significant effect.

%While we used a fixed $\gamma=10$ for this work, in future work we will explore a scheduling strategy to gradually increase $\gamma$ rather than have a fixed value in the future for potential further improvements. 

\begin{table}
    \centering
    \caption{Comparison between RL strategies and proposed approach on profit generation for 2021, 2022 and 2023. The best strategy from the proposed approach outperforms the RL results}
    \begin{tabular}{lcccl}
        \toprule
        \textbf{Models} & \textbf{2023 Profit\%} & \textbf{2022 Profit\%} & \textbf{2021 Profit\%}\\
        \midrule
         \textbf{\textit{A2C}}& 12.94 & -12.36 & 42.24 \\
         % \hline
         \textbf{\textit{PPO}}& 27.31 & 2.81 & 41.58 \\
         % \hline
         \textbf{\textit{DDPG}}& 33.75 & -17.09 & 39.79 \\
         % \hline
         \textbf{\textit{TD3}}& 40.96 & -17.63 & 41.55\\
         % \hline
         \textbf{\textit{SAC}}& 39.13 & -11.05 & 43.00\\
         % \hline
         \midrule
         \textbf{\textit{Buy and Hold}} & 27.77 & -20.64 & 36.05 \\
         \midrule
         \textbf{\textit{DeformTime + $Ret_t$ + L3 (Ours)}} & 38.98 & 21.61 &  36.64\\
         \textbf{\textit{DeformTime + $PRC_t$ + L3 (Ours)}} & {\bf 51.75} & 53.18 &  39.23\\

         \textbf{\textit{Crossformer + $RET_t$ + L3 (Ours)}} & 38.61 & 37.90 &  32.78\\
         \textbf{\textit{Crossformer + $PRC_t$ + L3 (Ours)}} & 51.42 & 51.04 &  48.62 \\

        \textbf{\textit{Autoformer + $Ret_t$ + L3 (Ours)}} & 1.62 & 4.12 &  3.21\\
         \textbf{\textit{Autoformer + $PRC_t$ + L3 (Ours)}} & -0.02 & 4.79 &  5.60\\
         \midrule

        \textbf{\textit{TimesNet + $Ret_t$ + L3 (Ours)}} & 17.27 & 15.89 &  20.77\\
         \textbf{\textit{TimesNet + $PRC_t$ + L3 (Ours)}} & 49.86 & 50.13 &  46.82\\   
         \midrule

         \textbf{\textit{DLinear + $Ret_t$ + L3 (Ours)}} & 14.63 & 29.03 &  43.72\\
         \textbf{\textit{DLinear + $PRC_t$ + L3 (Ours)}} & 39.11 & \textbf{53.91} &  \textbf{50.94}\\
         \midrule

         \textbf{\textit{P-sLSTM + $Ret_t$ + L3 (Ours)}} & 25.40 & 15.29 &  22.35\\
         \textbf{\textit{P-sLSTM + $PRC_t$ + L3 (Ours)}} & 25.89 & 12.05 &  8.48\\
         \bottomrule
    \end{tabular}
    \label{tab:rl-results}
\end{table}

As the the smoothed functions loss functions provided the best results, the smoothed versions were used for the other test periods, with results for the years $2021$, $2022$, and $2023$ being summarized in Table~\ref{tab:rl-results} for loss enabled models as well as with the selected RL strategies, \ie, DDPG, PPO, SAC, TD3 and A2C. For RL methods, we compared their performance on the stock features using the FinRL library. In addition, the profit generated by the baseline \texttt{Buy and Hold} strategy is also shown. In this strategy, a share is bought for each of the company stocks at the beginning of the testing period and sold at the end. The net profit generated by this strategy is essentially the difference between the stock prices at those corresponding times.

\noindent \textbf{Observations}: The results from Table~\ref{tab:rl-results} demonstrate that by simply updating the final output of potentially any artificial neural network, it can be made to learn how to generate strategies based on the proposed loss function(s). The results from these loss functions outperform the RL strategies as well, and in some cases quite significantly as in the bear market year of 2022. Furthermore, using \texttt{Buy and Hold} as the baseline strategy, nearly all of the RL models and our approach beat the baseline.

Moreover, observe that, with price ($PRC_t$) values for the loss functions, the profit margins are somewhat better than using return ($Ret_t$) from Tables~\ref{tab:rl-results}. This is surprising since the return is a better indicator of market direction and time, across stocks, rather than simply price; a possible explanation for this effect is that stock prices are non-stationary, unlike returns. As such, artificial neural networks are able to understand the nonlinear dynamics in market fluctuations better with respect to prices than returns and, thus, are able to find overall (useful) strategies. On performing the Mann-Whitney U-Test, it was observed that the results generated by the selected models are in fact statistically significant ($p < \alpha$) as compared to that generated by the RL strategies (refer Table \ref{tab:stat-test}). Interestingly, we also found that especially for the top performing models, including a hold node to mitigate risk did not have a significant effect on performance. We speculate that this may be because these models are predicting stock increases and decreases well enough that it does not make sense to hold any money, however this will require further investigation. These observations also help to answer our initial research questions.

\hangindent=3.6em %\hangafter=1
\textbf{RQ1.} \emph{Can a directly learned strategy allow any deep learning architecture for time-series forecasting to learn inter-stock patterns to maximize profits?~} - The results from Table~\ref{tab:rl-results} demonstrate that, by implementing and using our proposed loss functions, transformer models (or broadly speaking neural time-series models) can be used to directly learn and provide a trading strategy. Overall, the proposed method does indeed generate good profit margins. However in some cases, like in the Autoformer, or for the loss variants I and II, the model learns not to trade so as to avoid significant losses, which appears to result in a sub-optimal approach.

\hangindent=3.6em %\hangafter=1
\textbf{RQ2.} \emph{How would the performance of such a strategy compare across different TSF models as well as RL models?~} The DeformTime, Crossformer and DLinear models using PRC instead of RET all outperform the RL methods, and in some cases quite significantly. However, one model was not uniformly best across all test periods. It may be possible to further improve performance utilizing an ensemble strategy.

%- Comparing the performance of the proposed approach with the results, the best profit margin is reached by the DeformTime model trained on our proposed loss function variant III or \textit{StockLoss-L2}. It achieves a maximum profit of $51.75$\% while among the RL strategies TD3 achieves the highest profit of nearly $41$\% on the 2023 test data.

% \begin{figure*}[!h]
%     \centering
%     \includegraphics[scale=0.27]{images/stock_noExtra.png}
%     \caption{Plot denoting daily investment capital against daily profit/loss for specific run. The profit/loss follows an almost stationary pattern.}
%     \label{fig:captail-ret}
% \end{figure*}

% \begin{figure*}[!th]
%     \centering
%     \includegraphics[scale=0.12]{images/stock_noExtra_vi.png}
%     \caption{Variation of $\hat{V_i}$ against $O_i$ and $RET_t$ across the testing period for stock ticker \textbf{\textit{ATO}} for a particular run}
%     \label{fig:vi-ret}
% \end{figure*}

%% file: sections/07--conclusion.tex
\section{Conclusion}
\label{sec:conclusion}

In this paper, we introduce several novel loss functions that can essentially drive any time-series forecasting artificial neural network to directly learn a trading strategy while maximizing profits.  Our results demonstrated that this methodology outperformed other model variants/baselines as well as actor-critic algorithms. In particular, our Loss III \ie StockLoss-L2 was shown to provide a trading strategy with the highest maximum annual returns, which outperformed the best RL models by approximately 10\% in test periods 2021 and 2023, and notably achieved a return of 53.91\% in 2022, as compared to a 2.81\% return from the best performing RL model for that year. This demonstrates that, with a good artificial neural network capable of capturing the intricate features and pattern in stock data, simply changing the final output layer using our provided loss functions can lead to useful computational schemes for uncovering trading strategies to generate profit. A current limitation is the forecast are for an entire year, whereas in a production environment the models would have to trained every week or month. Future work will involve applying the proposed approach to neural architecture search-based methods, in order to facilitate the construction of custom neural models that identify/model the complex dependencies across the stock data.

%While experiments were conducted using transformer models, future work will involve applying the proposed approach to RNN models (and even neural architecture search-based methods) in order to facilitate the construction of custom networks that identify/model the complex dependencies across the stock data. %Furthermore, for the smoothness approximation a scheduling strategy for gradually increasing the value of $\gamma$ will be explored as opposed to a fixed value.

% \subsection*{Acknowledgments}
% This material is based upon work supported by the United States National Science Foundation under grant [redacted].

%\#2225354.

%% file: sections/08--appendix.tex
\section*{Technical Appendices and Supplementary Material}

\subsection*{Dataset Description}

The data was extracted from the Center for Research in Security Prices (CRSP)\footnote{https://www.crsp.org} which is a private data source and cannot be shared publicly. However, the data features that were selected have been highlighted in Table \ref{tab:financial_metrics}, and anyone with access to this dataset can prepare the data utilizing the code and instructions found at \url{https://anonymous.4open.science/r/data_processor-E183/README.md}.

\begin{table}[!h]
\caption{Economic Features Used for Stock Trading}
\label{tab:financial_metrics}
\centering
\begin{tabular}{lccl}
\toprule
\textbf{Features} & \textbf{Description} \\
\midrule
\emph{\textbf{Volume Change}} & Percentage change in trading volume \\
% \hline
\emph{\textbf{Bid-Ask Spread}} & \((AskPrice-BidPrice)/Price\) \\
% \hline
\emph{\textbf{Illiquidity}} & \(Return / (Volume \times Price)\) \\
% \hline
\emph{\textbf{Turn Over}} & \(Volume / SharesOutstanding\) \\
% \hline
\emph{\textbf{Price}} & Stock Price \\
% \hline
\emph{\textbf{Return}} & Percentage change in stock price \\
% \hline
% \emph{\textbf{DJI Return}} & DJI index return \\
% \emph{\textbf{S\&P 500 Return}} & S\&P 500 index return \\
% \hline
\emph{\textbf{Shares Outstanding}} & \textit{Authorised shares - Treasury stock} \\
% \hline
\emph{\textbf{Market Capitalization}} & \textit{Share Price $\times$ Shares Outstanding} \\
\bottomrule
\end{tabular}
\end{table}

% \subsection*{Experimental Results - Smooth vs Non-Smooth Loss Functions}

% The loss function formulations with the \texttt{sgn}() function were discontinuous and hence not perfectly differentiable. This affected the backpropagation and hance the ability of the neural networks to learn the finer temporal patterns for a particular stock and even between different stocks.

% \subsection*{Experimental Results - Statistical Significance between Hold and No-Hold}

% \begin{table}[!h]
% \caption{Mann-Whitney U-Test between Hold and No-Hold results for all models}
% \label{tab:man-u-w-models-hold}
% \centering
% \begin{tabular}{lcccccccl}
% \toprule
%  & \textbf{DeformTime} & \textbf{Crossformer} & \textbf{Autoformer} & \textbf{TimesNet} & \textbf{DLinear} & \textbf{P-sLSTM} \\
% \midrule
% \emph{p-values} & 0.47 & 0.31 & 0.91 & 0.47 & & 0.57 \\
% \bottomrule
% \end{tabular}
% \end{table}

\subsection*{Experimental Results - Testing Years and Portfolios}
The models were tested on 3 different testing years to note the profit generation capacity. These were 2021 , 2022 and 2023. Tables \ref{tab:results-ret-2023} and \ref{tab:results-prc-2023} highlight the testing results for the year 2023. Similarly Tables \ref{tab:results-ret-2022} and \ref{tab:results-prc-2022} shows the testing results for 2022 and, Tables \ref{tab:results-ret-2021} and \ref{tab:results-prc-2021} highlight the testing results for the year 2021. DLinear performed the overall best with the third loss function variant achieving an annual profit of 53.91\% followed by DeformTime (53.18\%) using the $PRC_t$ Loss III variant for 2022. For 2021, DLinear performed the best with an annual profit of 50.94\% followed by TimesNet with a profit of 46.82\%. 

Also, the results with the loss functions are overall higher than any of the RL strategies for each of the testing years, and even the \texttt{Buy \& Hold} strategy (Refer Table \ref{tab:rl-results}).

Apart from the original portfolio of 50 companies, a different medium capitalization portfolio ( where market capitalization is between \$2 Billion and \$10 Billion) consisting of 30 companies was used to ascertain the robustness of the loss functions across varying portfolios. The results for this are available in Tables \ref{tab:results-ret-m1} and \ref{tab:results-prc-m1}. With $Ret_t$ P-sLSTM performed best loss L3 with a profit of 39.57\%. DeformTime performed the best on L3 with the hold and $PRC_t$ with a profit of 27.03\%.

Statistical significance for the models with the L3 loss function were also compared to the different RL models for 2023 as the test period, as shown in Table~\ref{tab:stat-test}. These results show that the models adapted with our loss function had statistically significant differences from the RL models, which in most cases had the adapted models outperforming the RL models.

% Tables \ref{tab:results-ret-2021} and \ref{tab:results-prc-2021} highlight the testing results for the year 2021. Again, DLinear performed the best with an annual profit of 51.21\% followed by TimesNet with a profit of 46.82\%. This in turn was also higher than any of the RL strategies and even the \texttt{Buy \& Hold} strategy for the same year (Refer Table \ref{tab:rl-results}).

\begin{table*}[!th]
    \centering
    \caption{Comparison of the highest profit percentages generated by different transformer based models for the proposed smooth loss functions with \textit{RET} for 2023}
    \begin{tabular}{lccccc|cccccl}
    \toprule
    \multirow{3}{*}{\textbf{\makecell{Input Seq.\\ Length}}} & \multirow{3}{*}{\textbf{Model}} & \multicolumn{8}{c}{\textbf{Profit \%}} \\
    \cmidrule(lr){3-10}
    & & \multicolumn{4}{c}{\textbf{Hold}} & \multicolumn{4}{c}{\textbf{No Hold}}\\
    \cmidrule(lr){3-10}
    & & \textbf{L-1} & \textbf{L-2} & \textbf{L-3} & \textbf{L-4} & \textbf{L-1} & \textbf{L-2} & \textbf{L-3} & \textbf{L-4} \\
    \midrule

    \multirow{3}{*}{\textbf{\textit{96}}} & \textbf{\textit{DeformTime}} & 0 & 0 & \textbf{38.98} & 5.19 & 7.13 & 9.36 & 21.57 & 3.15 \\
    % \cmidrule(lr){2-10}
    & \textbf{\textit{Crossformer}} & 2.57 & 0 & 38.61 & 6.59 & 7.87 & -1.73 & -0.05 & 9.19 \\
    % \cmidrule(lr){2-10}
    & \textbf{\textit{AutoFormer}} & 0.07 & 0.0 & 1.62 & 0.14 & 0.15 & 1.52 & 0.42 & 0.37 \\
    & \textbf{\textit{TimesNet}} & 0.19 & 0.0 & 2.79 & 8.85 & 6.18 & 5.54 & 17.27 & 2.40 \\
    & \textbf{\textit{DLinear}} & 5.15 & 0.0 & 14.63 & 3.99 & 9.73 & 6.90 & 10.28 & 2.92 \\
    & \textbf{\textit{P-sLSTM}} & 23.45 & 6.28 & 25.40 & 15.36 & 11.52 & -0.46 & 15.37 & 5.45 \\
    \bottomrule
    \end{tabular}
    \label{tab:results-ret-2023}
\end{table*}

\begin{table*}[!h]
    \centering
    \caption{Comparison of the highest profit percentages generated from different transformer models for the proposed smooth loss functions with \textit{PRC} for 2023}
    \begin{tabular}{lccccc|cccccl}
    \toprule
    \multirow{3}{*}{\textbf{\makecell{Input Seq.\\ Length}}} & \multirow{3}{*}{\textbf{Model}} & \multicolumn{8}{c}{\textbf{Profit \%}} \\
    \cmidrule(lr){3-10}
    & & \multicolumn{4}{c}{\textbf{Hold}} & \multicolumn{4}{c}{\textbf{No Hold}}\\
    \cmidrule(lr){3-10}
    & & \textbf{L-1} & \textbf{L-2} & \textbf{L-3} & \textbf{L-4} & \textbf{L-1} & \textbf{L-2} & \textbf{L-3} & \textbf{L-4} \\
    \midrule

    \multirow{6}{*}{\textbf{\textit{96}}} & \textbf{\textit{DeformTime}} & 0 & 0 & 51.57 & 7.01 & 29.13 & 9.36 & \textbf{51.75} & 6.70 \\
    % \cmidrule(lr){2-10}
    & \textbf{\textit{Crossformer}} & 3.19 & 13.58 & 51.42 & 9.51 & 2.61 & 8.87 & 51.35 & 10.04 \\
    % \cmidrule(lr){2-10}
    & \textbf{\textit{AutoFormer}} & 0.23 & 0.0 & -0.02 & 1.7 & 0.07 & 1.13 & -0.006 & -0.03 \\
    & \textbf{\textit{TimesNet}} & 7.86 & 0.0 & 49.31 & 8.28 & 5.36 & 4.91 & 49.86 & 9.72 \\
    & \textbf{\textit{DLinear}} & 0.98 & 0.0 & 34.72 & 7.38 & 7.64 & 6.11 & 39.11 & 6.22 \\
    & \textbf{\textit{P-sLSTM}} & 7.94 & 2.53 & 15.91 & 8.56 & 9.16 & 8.97 & 25.89 & 5.21 \\
    \bottomrule
    \end{tabular}
    \label{tab:results-prc-2023}
\end{table*}

\begin{table*}[!h]
    \centering
    \caption{Comparison of the highest profit percentages generated from different transformer models for the proposed smooth loss functions with \textit{RET} for 2022}
    \begin{tabular}{lccccc|cccccl}
    \toprule
    \multirow{3}{*}{\textbf{\makecell{Input Seq.\\ Length}}} & \multirow{3}{*}{\textbf{Model}} & \multicolumn{8}{c}{\textbf{Profit \%}} \\
    \cmidrule(lr){3-10}
    & & \multicolumn{4}{c}{\textbf{Hold}} & \multicolumn{4}{c}{\textbf{No Hold}}\\
    \cmidrule(lr){3-10}
    & & \textbf{L-1} & \textbf{L-2} & \textbf{L-3} & \textbf{L-4} & \textbf{L-1} & \textbf{L-2} & \textbf{L-3} & \textbf{L-4} \\
    \midrule

    \multirow{6}{*}{\textbf{\textit{96}}} & \textbf{\textit{DeformTime}} & 0.0 & 0.0 & -0.38 & 1.67 & 7.60 & 1.04 & 21.61 & 0.27 \\
    % \cmidrule(lr){2-10}
    & \textbf{\textit{Crossformer}} & 0.93 & 0.0 & 2.01 & 1.96 & 7.74 & 2.79 & \textbf{37.90} & 5.62 \\
    % \cmidrule(lr){2-10}
    & \textbf{\textit{AutoFormer}} & 1.11 & 1.05 & 0.43 & 1.7 & 6.39 & 5.12 & 4.12 & 5.29\\
    & \textbf{\textit{TimesNet}} & 0.0 & 0.46 & 15.89 & 9.58 & 5.83 & 9.51 & 10.99 & 8.95 \\
    & \textbf{\textit{DLinear}} & 0.0 & 1.69 & 29.03 & 6.82 & 6.46 & 6.29 & 10.06 & 5.91 \\
    & \textbf{\textit{P-sLSTM}} & 15.99 & 3.18 & 15.29 & 10.58 & 1.46 & -0.81 & 13.19 & 11.72 \\
    \bottomrule
    \end{tabular}
    \label{tab:results-ret-2022}
\end{table*}

\begin{table*}[!h]
    \centering
    \caption{Comparison of the highest profit percentages generated from different transformer models for the proposed smooth loss functions with \textit{PRC} for 2022}
    \begin{tabular}{lccccc|cccccl}
    \toprule
    \multirow{3}{*}{\textbf{\makecell{Input Seq.\\ Length}}} & \multirow{3}{*}{\textbf{Model}} & \multicolumn{8}{c}{\textbf{Profit \%}} \\
    \cmidrule(lr){3-10}
    & & \multicolumn{4}{c}{\textbf{Hold}} & \multicolumn{4}{c}{\textbf{No Hold}}\\
    \cmidrule(lr){3-10}
    & & \textbf{L-1} & \textbf{L-2} & \textbf{L-3} & \textbf{L-4} & \textbf{L-1} & \textbf{L-2} & \textbf{L-3} & \textbf{L-4} \\
    \midrule

    \multirow{6}{*}{\textbf{\textit{96}}} & \textbf{\textit{DeformTime}} & 0.22 & 0.0 & -0.03 & 2.44 & 30.55 & 19.34 & 53.18 & 15.92 \\
    % \cmidrule(lr){2-10}
    & \textbf{\textit{Crossformer}} & 2.37 & 1.59 & -0.13 & 2.81 & 7.40 & 9.51 & 51.04 & 12.68 \\
    % \cmidrule(lr){2-10}
    & \textbf{\textit{AutoFormer}} & 1.51 & 0.68 & 0.76 & 0.68 & 5.92 & 4.08 & 4.79 & 5.23 \\
    & \textbf{\textit{TimesNet}} & 12.11 & 0.0 & 49.31 & 9.58 & 13.37 & 12.68 & 50.13 & 9.24 \\
    & \textbf{\textit{DLinear}} & 4.59 & 3.38 & 0.0 & 4.59 & 2.99 & 4.91 & \textbf{53.91} & 6.22 \\
    & \textbf{\textit{P-sLSTM}} & 10.25 & 9.09 & 12.05 & 10.25 & 8.76 & 9.01 & 9.40 & 8.14 \\
    \bottomrule
    \end{tabular}
    \label{tab:results-prc-2022}
\end{table*}

\begin{table*}[!t]
    \centering
    \caption{Comparison of the highest profit percentages generated from different transformer models for the proposed smooth loss functions with \textit{RET} for 2021 }
    \begin{tabular}{lccccc|cccccl}
    \toprule
    \multirow{3}{*}{\textbf{\makecell{Input Seq.\\ Length}}} & \multirow{3}{*}{\textbf{Model}} & \multicolumn{8}{c}{\textbf{Profit \%}} \\
    \cmidrule(lr){3-10}
    & & \multicolumn{4}{c}{\textbf{Hold}} & \multicolumn{4}{c}{\textbf{No Hold}}\\
    \cmidrule(lr){3-10}
    & & \textbf{L-1} & \textbf{L-2} & \textbf{L-3} & \textbf{L-4} & \textbf{L-1} & \textbf{L-2} & \textbf{L-3} & \textbf{L-4} \\
    \midrule

    \multirow{6}{*}{\textbf{\textit{96}}} & \textbf{\textit{DeformTime}} & 0.0 & 0.0 & 5.43 & 0.60 & 7.98 & 17.86 & 36.64 & 2.32 \\
    % \cmidrule(lr){2-10}
    & \textbf{\textit{Crossformer}} & 0.83 & 0.40 & 1.16 & 1.11 & 5.97 & 0.75 & 32.78 & 8.75\\
    % \cmidrule(lr){2-10}
    & \textbf{\textit{AutoFormer}} & 0.95 & 0.43 & -0.22 & 0.67 & 4.51 & 3.16 & 3.21 & 2.03 \\
    & \textbf{\textit{TimesNet}} & 1.91 & 0.0 & 8.05 & 4.15 & 10.94 & 9.27 & 20.77 & 8.23 \\
    & \textbf{\textit{DLinear}} & 4.78 & 5.58 & 5.41 & 3.63 & 7.49 & 5.85 & \textbf{43.72} & 4.03 \\
    & \textbf{\textit{P-sLSTM}} & 12.88 & 7.63 & 22.35 & 1.37 & 18.21 & 9.17 & 17.09 & -1.81 \\
    \bottomrule
    \end{tabular}
    \label{tab:results-ret-2021}
\end{table*}

\begin{table*}[!h]
    \centering
    \caption{Comparison of the highest profit percentages generated from different transformer models for the proposed smooth loss functions with \textit{PRC} for 2021}
    \begin{tabular}{lccccc|cccccl}
    \toprule
    \multirow{3}{*}{\textbf{\makecell{Input Seq.\\ Length}}} & \multirow{3}{*}{\textbf{Model}} & \multicolumn{8}{c}{\textbf{Profit \%}} \\
    \cmidrule(lr){3-10}
    & & \multicolumn{4}{c}{\textbf{Hold}} & \multicolumn{4}{c}{\textbf{No Hold}}\\
    \cmidrule(lr){3-10}
    & & \textbf{L-1} & \textbf{L-2} & \textbf{L-3} & \textbf{L-4} & \textbf{L-1} & \textbf{L-2} & \textbf{L-3} & \textbf{L-4} \\
    \midrule

    \multirow{6}{*}{\textbf{\textit{96}}} & \textbf{\textit{DeformTime}} & 0.0 & 0.0 & 5.34 & 1.67 & 2.75 & 3.85 & 39.23 & 5.61 \\
    % \cmidrule(lr){2-10}
    & \textbf{\textit{Crossformer}} & 1.16 & 1.35 & 0.65 & 1.38 & 7.03 & 8.36 & 48.62 & 8.42 \\
    % \cmidrule(lr){2-10}
    & \textbf{\textit{AutoFormer}} & 0.48 & 0.42 & 0.67 & 0.69 & 3.52 & 3.05 & 5.60 & 2.17 \\
    & \textbf{\textit{TimesNet}} & 5.67 & 0.0 & 46.58 & 8.74 & 7.54 & 9.78 & 46.82 & 5.17 \\
    & \textbf{\textit{DLinear}} & 3.90 & 0.0 & \textbf{50.94} & 4.81 & 5.42 & 5.53 & 4.18 & 2.90 \\
    & \textbf{\textit{P-sLSTM}} & 11.39 & 10.56 & 8.48 & 8.89 & 7.16 & 8.10 & 8.15 & -7.17 \\
    \bottomrule
    \end{tabular}
    \label{tab:results-prc-2021}
\end{table*}

% \begin{table}[!h]
%     \centering
%     \caption{Comparison between RL strategies and proposed approach on profit generation for 2022 and 2021. The best strategy from the proposed approach outperforms the RL results}
%     \begin{tabular}{lcccl}
%         \toprule
%         \textbf{Models} & \textbf{2022 Profit\% (S\&P features)} & \textbf{2021 Profit\% (S\&P features)}\\
%         \midrule
%          \textbf{\textit{A2C}} & -12.36 & 42.24 \\
%          % \hline
%          \textbf{\textit{PPO}} & 2.81 & 41.58 \\
%          % \hline
%          \textbf{\textit{DDPG}} & -17.09 & 39.79 \\
%          % \hline
%          \textbf{\textit{TD3}} & -17.63 & 41.55\\
%          % \hline
%          \textbf{\textit{SAC}} & -11.05 & 43.00\\
%          % \hline
%          \midrule
%          \textbf{\textit{Buy and Hold}} & -20.64 & 36.05 \\
%          \midrule
%          \textbf{\textit{$PRC_t$ + Loss III (Ours)}} & 53.91 (DLinear) &  51.21 (DLinear)\\
%          \textbf{\textit{$RET_t$ + Loss III (Ours)}} & 37.90 (Crossformer) &  43.72 (DLinear)\\
%          % \hline
%          \bottomrule
%     \end{tabular}
%     \label{tab:rl-results-2021-2022}
% \end{table}

\begin{table*}[!h]
    \centering
    \caption{Comparison of the highest profit percentages generated from different models for the proposed smooth loss functions with \textit{RET} for 2023 for the new portfolio}
    \begin{tabular}{lccccccccccl}
    \toprule
    \multirow{3}{*}{\textbf{\makecell{Input Seq.\\ Length}}} & \multirow{3}{*}{\textbf{Model}} & \multicolumn{8}{c}{\textbf{Profit \%}} \\
    \cmidrule(lr){3-10}
    & & \multicolumn{4}{c}{\textbf{Hold}} & \multicolumn{4}{c}{\textbf{No Hold}}\\
    \cmidrule(lr){3-10}
    & & \textbf{L-1} & \textbf{L-2} & \textbf{L-3} & \textbf{L-4} & \textbf{L-1} & \textbf{L-2} & \textbf{L-3} & \textbf{L-4} \\
    \midrule

    \multirow{6}{*}{\textbf{\textit{96}}} & \textbf{\textit{DeformTime}} & 0.01 & 0.02 & 2.54 & 2.18 & 14.18 & 6.06 & 27.14 & 7.59 \\
    % \cmidrule(lr){2-10}
    & \textbf{\textit{Crossformer}} & 0.007 & 0.23 & 3.41 & 1.96 & -7.15 & -9.77 & 31.91 & 6.61 \\
    % \cmidrule(lr){2-10}
    & \textbf{\textit{AutoFormer}} & 9.81 & 7.26 & 9.26 & 5.85 & 7.74 & 8.92 & 7.34 & 5.63\\
    & \textbf{\textit{TimesNet}} & 10.25 & 7.62 & 33.19 & 8.64 & 8.14 & 19.89 & 9.31 & 19.56 \\
    & \textbf{\textit{DLinear}} & 8.01 & 8.03 & 7.51 & 4.98 & 2.32 & 6.19 & 7.83 & 5.99 \\
    & \textbf{\textit{P-sLSTM}} & 9.29 & 2.51 & \textbf{39.57} & 9.13 & 28.20 & 15.11 & 21.90 & 21.48 \\
    \bottomrule
    \end{tabular}
    \label{tab:results-ret-m1}
\end{table*}

\begin{table*}[!h]
    \centering
    \caption{Comparison of the highest profit percentages generated from different models for the proposed smooth loss functions with \textit{PRC} for 2023 for the new portfolio}
    \begin{tabular}{lccccccccccl}
    \toprule
    \multirow{3}{*}{\textbf{\makecell{Input Seq.\\ Length}}} & \multirow{3}{*}{\textbf{Model}} & \multicolumn{8}{c}{\textbf{Profit \%}} \\
    \cmidrule(lr){3-10}
    & & \multicolumn{4}{c}{\textbf{Hold}} & \multicolumn{4}{c}{\textbf{No Hold}}\\
    \cmidrule(lr){3-10}
    & & \textbf{L-1} & \textbf{L-2} & \textbf{L-3} & \textbf{L-4} & \textbf{L-1} & \textbf{L-2} & \textbf{L-3} & \textbf{L-4} \\
    \midrule

    \multirow{6}{*}{\textbf{\textit{96}}} & \textbf{\textit{DeformTime}} & 2.99 & 0.27 & \textbf{27.03} & 7.35 & 7.29 & 9.55 & 25.63 & 11.44 \\
    % \cmidrule(lr){2-10}
    & \textbf{\textit{Crossformer}} & 5.26 & 1.69 & 25.60 & 7.26 & 7.24 & 4.81 & 26.61 & 9.06 \\
    % \cmidrule(lr){2-10}
    & \textbf{\textit{AutoFormer}} & 6.37 & 3.80 & 6.14 & 3.22 & 9.79 & 4.28 & 4.04 & 3.92 \\
    & \textbf{\textit{TimesNet}} & 11.87 & 5.59 & 15.54 & 9.17 & 8.65 & 19.52 & 8.88 & 12.89 \\
    & \textbf{\textit{DLinear}} & 8.11 & 7.26 & 6.32 & 4.87 & 2.75 & 6.09 & 7.58 & 4.61 \\
    & \textbf{\textit{P-sLSTM}} & 22.79 & 15.32 & 4.27 & 12.13 & 13.13 & 12.03 & 12.74 & 11.02 \\
    \bottomrule
    \end{tabular}
    \label{tab:results-prc-m1}
\end{table*}

\begin{table}[!h]
    \centering
    \caption{Mann-Whitney U-Test $p-$values between No-Hold Sequence Models for Loss L3 and RL Models at confidence level $\alpha=0.05$}
    \begin{tabular}{lccccccl}
        \toprule
        \textbf{Models} & \textbf{A2C} & \textbf{PPO} & \textbf{DDPG} & \textbf{SAC} & \textbf{TD3}\\
        \midrule
        \textbf{\textit{DeformTime}}& p=0.00 & p=0.01 & p=0.00 & p=0.00 & p=0.01 \\
        \textbf{\textit{Crossformer}}& p=0.02 & p=0.00 & p=0.00 & p=0.01 & p=0.00 \\
        \textbf{\textit{Autoformer}}& p=0.00 & p=0.00 & p=0.03 & p=0.00 & p=0.02 \\
        \textbf{\textit{TimesNet}}& p=0.01 & p=0.01 & p=0.00 & p=0.03 & p=0.00\\
        \textbf{\textit{DLinear}}& p=0.00 & p=0.02 & p=0.00 & p=0.01 & p=0.00 \\
        \textbf{\textit{P-sLSTM}}& p=0.01 & p=0.00 & p=0.01 & p=0.00 & p=0.00 \\
         \bottomrule
    \end{tabular}
    \label{tab:stat-test}
\end{table}